\documentclass[10pt,twocolumn,letterpaper]{article}

\usepackage[pagenumbers]{cvpr} 

\usepackage{graphicx}
\usepackage{amsmath}
\usepackage{amssymb}
\usepackage{booktabs}
\usepackage{multirow}
\usepackage{color}
\usepackage{soul}
\usepackage{pifont}
\usepackage{threeparttable}
\usepackage[pagebackref,breaklinks,colorlinks]{hyperref}

\usepackage[capitalize]{cleveref}
\crefname{section}{Sec.}{Secs.}
\Crefname{section}{Section}{Sections}
\Crefname{table}{Table}{Tables}
\crefname{table}{Tab.}{Tabs.}

\def\cvprPaperID{2942} 
\def\confName{CVPR}
\def\confYear{2022}

\usepackage{amsmath,amsfonts,bm}

\def\eqref#1{equation~\ref{#1}}

\def\1{\bm{1}}

\def\rvc{{\mathbf{c}}}
\def\rvd{{\mathbf{d}}}

\def\rvo{{\mathbf{o}}}

\def\rvr{{\mathbf{r}}}

\def\rvx{{\mathbf{x}}}

\def\rmR{{\mathbf{R}}}
\def\rmS{{\mathbf{S}}}

\DeclareMathAlphabet{\mathsfit}{\encodingdefault}{\sfdefault}{m}{sl}
\SetMathAlphabet{\mathsfit}{bold}{\encodingdefault}{\sfdefault}{bx}{n}

\def\name{NeuSample\@\xspace}

\begin{document}

\title{\name: Neural Sample Field for Efficient View Synthesis}

\author{
Jiemin Fang$^{1,2}$ ,  Lingxi Xie$^{3}$ ,  Xinggang Wang$^2$\footnotemark[2] ,  Xiaopeng Zhang$^{3}$ ,  Wenyu Liu$^2$ ,  Qi Tian$^{3}$\\
$^1$Institute of Artificial Intelligence, Huazhong University of Science \& Technology\\
$^2$School of EIC, Huazhong University of Science \& Technology \; $^3$Huawei Inc.\\
\texttt{\small\{jaminfong, xgwang, liuwy\}@hust.edu.cn}\\
\texttt{\small\{198808xc, zxphistory\}@gmail.com} \; \texttt{\small tian.qi1@huawei.com}
\\
\small \href{https://jaminfong.cn/neusample/}{\texttt{jaminfong.cn/neusample/}}
}
\maketitle

\begin{abstract}
Neural radiance fields (NeRF) have shown great potentials in representing 3D scenes and synthesizing novel views, but the computational overhead of NeRF at the inference stage is still heavy. To alleviate the burden, we delve into the coarse-to-fine, hierarchical sampling procedure of NeRF and point out that the coarse stage can be replaced by a lightweight module which we name a neural sample field. The proposed sample field maps rays into sample distributions, which can be transformed into point coordinates and fed into radiance fields for volume rendering. The overall framework is named as \name. We perform experiments on Realistic Synthetic 360$^{\circ}$ and Real Forward-Facing, two popular 3D scene sets, and show that \name achieves better rendering quality than NeRF while enjoying a faster inference speed. \name is further compressed with a proposed sample field extraction method towards a better trade-off between quality and speed. 
\end{abstract}
{
\renewcommand{\thefootnote}{\fnsymbol{footnote}}
\footnotetext[2]{Corresponding author.}}


\section{Introduction}
\label{sec:intro}
Novel view synthesis is a long-standing and important problem in computer vision~\cite{carranza2003free,szeliski2019computer}, aiming at reconstructing a 3D object/scene with sparsely sampled views and generating images from unseen views. It has a wide range of applications, such as rendering interactive objects in virtual reality and offering 3D preview of objects/scenes.

Recently, researchers propose the concept of neural radiance fields (NeRF)~\cite{mildenhall2020nerf} that represent a scene with a continuous 5D function, which is often formulated using a deep neural network and hence can be optimized via gradient descent. The function takes point coordinates (3D) and the observer's view direction (2D) as input and outputs the pixel radiance and opacity. To connect real 3D points with image pixels, a classical technique volume rendering~\cite{kajiya1984ray} is used. It samples a number of points along the ray (starting from the observer, extending along the view direction), computes their density (\textit{a.k.a.} opacity) values and color properties, and eventually accumulate them into the final output. Compared to conventional methods (\textit{i.e.}, voxel-based or mesh-based representations), NeRF saves considerable storage by compressing each scene/object into a neural network.

\begin{figure}[!t]
\centering
\includegraphics[width=\columnwidth]{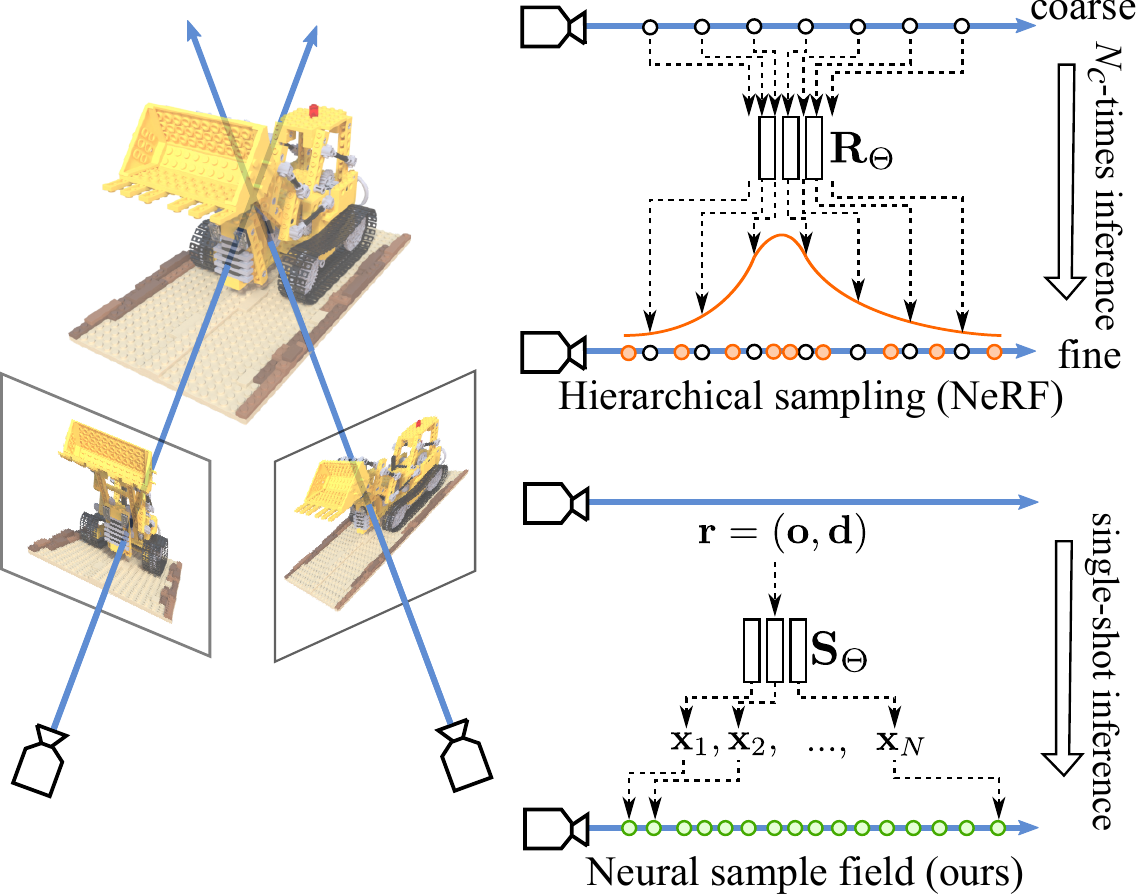}
\vspace{-15pt}
\caption{Comparisons between the conventional hierarchical sampling method and our \name. Previous hierarchical sampling needs to first infer a set of coarse points. Based on opacity properties of coarse ones, fine samples are obtained for elaborate rendering. \name directly maps a ray $\rvr = (\rvo, \rvd)$ into a sampling distribution with single inference by constructing a sample field. The obtained samples can directly render high-quality images.}
\vspace{-15pt}
\label{fig: sample_comp}
\end{figure}

To guarantee high rendering quality, NeRF and most subsequent methods~\cite{mildenhall2020nerf,Barron_2021_ICCV,Hedman_2021_ICCV,Reiser_2021_ICCV} adopted a hierarchical sampling method. As shown in Fig.~\ref{fig: sample_comp}, the algorithm first uniformly samples a set of $N_c$ (usually taken as $64$) coarse-level points along the ray, and feeds them into the radiance field to obtain the densities and color properties of these points, as well as an updated distribution to sample more fine-level points (for another round of inference). Finally, the outputs of coarse-level and fine-level points are accumulated as the final output. Although the strategy leads to improved rendering performance, the two-stage inference incurs heavy computational overheads, making it difficult to apply NeRF to real-life or interactive scenarios.

To alleviate the burden, we propose a novel framework named \textbf{\name} that only requires \textbf{single-shot} inference for sampling -- in other words, we discard the costly coarse stage and instead use a \textbf{neural sample field}, which takes a ray representation as input and outputs a sampling distribution along the ray. As shown in Fig.~\ref{fig: sample_comp}, the sample field is also a 5D function and formulated using a neural network, which takes the observer's coordinate $(x_o, y_o, z_o)$ and view direction $(\theta, \phi)$ as input and outputs $N$ numbers, corresponding to the distances between the points-to-sample and the observer. These points are then fed into the neural radiance field to accomplish the volume rendering procedure.

We perform experiments on two commonly used benchmarks, namely, Realistic Synthetic 360$^\circ$ and Real Forward-Facing. Compared to NeRF, \name demonstrates superior rendering quality and saves around a quarter of inference time. In addition, a sample field extraction method assists \name to achieve competitive rendering quality with $25\%$ of inference time. Diagnostic studies show that the improved quality-complexity trade-off owes to the neural sample field that learns an efficient way of sampling points. The contribution of this work is summarized as follows.
\begin{itemize}
    \vspace{-5pt}
    \item We propose a sample field to map one ray into a sample distribution. This field is parameterized as a neural network with fully connected layers.
    \vspace{-5pt}
    \item The proposed sample field can be integrated with radiance fields to perform volume rendering, which not only saves computation cost from coarse networks used in the conventional hierarchical sampling strategy but also shows stronger rendering quality than NeRF.
    \vspace{-5pt}
    \item A sample field extraction method is proposed to reduce the number of sampled points. This method further accelerates rendering while maintaining competitive rendering quality.
\end{itemize}

\section{Related Work}
\label{sec:relate}

\paragraph{Neural Implicit Representations}
Using neural representations to modelling 3D structures or geometry~\cite{atzmon2019controlling,genova2019learning,mescheder2019occupancy,michalkiewicz2019implicit,niemeyer2019occupancy,peng2020convolutional,liu2020neural,sitzmann2019scene,srinivasan2021nerv} has shown great success. Neural representations model 3D scenes in a continuous space and can be optimized with a differentiable manner. The storage can be saved with network inference. Most of above methods require explicit supervision. NeRF~\cite{mildenhall2020nerf} and subsequent works~\cite{schwarz2020graf,Barron_2021_ICCV,kaizhang2020,Park_2021_ICCV,pumarola2020d,yu2021pixelnerf,Jain_2021_ICCV} use neural radiance fields to map 3D coordinates into color and opacity values, which achieve strong performance on synthesizing photo-realistic images from novel views. Besides neural radiance fields, some works adopt other types of neural fields to represent or model 3D scenes, \eg textured material~\cite{henzler2020learning,oechsle2019texture,rainer2019neural,Rainer2020Unified}, indirect illumination values~\cite{ren2013global}, surfacing reconstruction~\cite{wang2021neus,Oechsle2021ICCV} and light fields~\cite{sitzmann2021lfns}. 
Our work is inspired by the neural field concept and proposes a sample field, which maps rays into sample distributions. This field is efficient and the predicted samples help radiance fields render high-quality images.

\vspace{-15pt}
\paragraph{Accelerating Rendering with Caching}
One main stream of methods accelerate rendering by pre-computing and storing explicit data structures. For inference, properties of points can be looked up and latency of neural network inference can be drastically reduced. Neural Sparse Voxel Fields (NSVF)~\cite{LiuGLCT20} allows for empty space skipping and early ray termination by constructing an octree structure. KiloNeRF~\cite{Reiser_2021_ICCV} represents a scene with thousands of tiny MLPs, each of which is only responsible for a cell in the space grid. SNeRG~\cite{Hedman_2021_ICCV}, FastNeRF~\cite{Garbin_2021_ICCV} and PlenOctrees~\cite{Yu_2021_ICCV} \etal adopt a similar methodology by pre-computing properties/features with sparse voxel grid-like structures using a learned NeRF. Directly querying these properties during inference can effectively save cost. NeX~\cite{wizadwongsa2021nex} represents scenes based on the multiplane image (MPI) with MPLs. The MPI grid can be cached for fast rendering. Most of the above methods achieve real-time rendering speed. This methodology can be taken as using storage to complement inference. Though storage may be compressed via techniques like quantization, for high resolution, storage is still a non-negligible element. From a different perspective, our method aims at improving rendering speed without caching and additional storage. \name is a parallel work to the above ones and can be integrated with caching-aided frameworks.

\vspace{-15pt}
\paragraph{Accelerating Rendering from Inference Times}
A series of methods promote the rendering efficiency by focusing on the neural representation itself and reducing the inference times. AutoInt~\cite{lindell2021autoint} proposes to automatically integrate output colors for rendering with neural networks, which approximates the integral along rays in piecewise sections. DONeRF~\cite{neff2021donerf} uses the depth information to guide sampling by training a ``depth oracle'' network, which greatly reduces the sample number for rendering. However, depth information is usually hard to obtain in real scenarios. Light Filed Networks~\cite{sitzmann2021lfns} directly map a ray into the color, only requiring one single inference for rendering rather than mapping hundreds of points and significantly promoting efficiency. This method discards explicit 3D point modelling so additional supervision may be needed to construct multi-view consistency towards complicated geometry. NeRF-ID~\cite{arandjelovic2021nerf} reduces the sample number by training an importance predictor, where a proposal network with Transformer/MLP-Mixer architectures is used for sampling based on coarse network features. TermiNeRF~\cite{piala2021terminerf} predicts weights of ray segments and further sample points based on the estimated weights. Our work maintains advantage of volume rendering for modelling multi-view consistency with explicit 3D point coordinates, but saves computation cost from cumbersome coarse fields. Samples are directly obtained by the proposed sample field.

\section{Method}
\label{sec:method}
In this section, we first review formulations in NeRF~\cite{mildenhall2020nerf}. Second, we introduce the proposed neural sample field and the overall framework for rendering. We finally introduce a sample field extraction method to produce fewer samples for further acceleration.

\subsection{Review of NeRF}
\paragraph{Field Construction and Optimization}
The neural radiance field is first proposed in \cite{mildenhall2020nerf}, which is designed to represent a scene with emitted radiance for each position in the space. The field is constructed by mapping the coordinate of a point into its volume density and color with neural networks. Denoting a point $\rvx$ with a 5D-coordinate $(x, y, z, \theta, \phi)$,  we predict its volume density and viewing color via neural networks as
\begin{equation}
\label{eq: field}
    \sigma, \rvc = \rmR_{\Theta} (x, y, z, \theta, \phi),
\end{equation}
where $x, y, z$ indicate the point location, $\theta, \phi$ are the view direction, and $\rmR_{\Theta}$ is the radiance field instantiated as a neural network.
To connect these points in the real space and images taken from the camera, one pixel in an image can be rendered by computing densities and colors of points along the according ray and performing the classic volume rendering~\cite{kajiya1984ray}. Specifically, the expected color $\hat{C}(\rvr)$ of the pixel along camera ray $\rvr(t) = \rvo + t\rvd$ is computed with the quadrature rule as: 
\vspace{-10pt}
\begin{equation}
\label{eq: render}
\begin{aligned}
    \hat{C}(\rvr) &= \sum_{i=1}^N T_i(1 - \text{exp}(-\sigma_i \delta_i))\rvc(\rvx_i, \rvd),\\
    T_i &= \text{exp}(- \sum_{j=1}^{i-1} \sigma (\rvx_j) \delta_j),
\end{aligned}
\vspace{-8pt}
\end{equation}
where $\rvo, \rvd$ denote the origin and direction of the ray respectively, $t$ denotes the distance of the sample from the origin, and $\delta_i$ is defined as the distance between two adjacent samples, \ie, $\delta_i = t_{i+1} - t_i$. The field defined in Eq.~\ref{eq: field} can be optimized with a differentiable manner by minimizing the loss between the rendered and ground truth color in the image:
\begin{equation}
    \label{eq: color_loss}
    \mathcal{L} = \lVert\hat{C}(\rvr) - C(\rvr)\rVert^2_2.
\end{equation}

\begin{figure*}[thbp]
    \centering
    \includegraphics[width=0.9\linewidth]{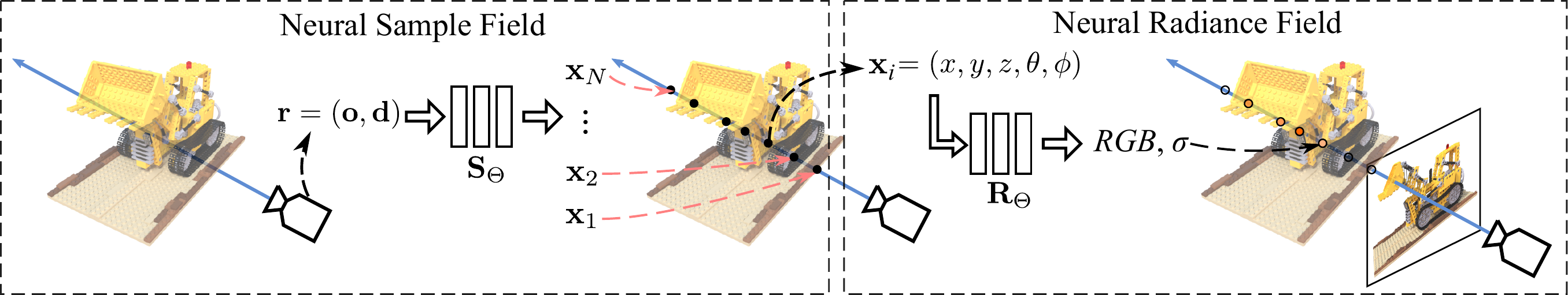}
    \caption{Overall framework of \name. To render a pixel in the image, the ray passing through the pixel is first fed into a sample field network $\rmS_{\Theta}$, which is mapped to a distribution along the ray. The distribution is then transformed to 3D-point coordinates, which are fed into the radiance field $\rmR_{\Theta}$ to obtain colors and densities. Finally, volume rendering is performed on these points.}
    \label{fig: framework}
\end{figure*}

\vspace{-15pt}
\paragraph{Volume Samling in Radiance Fields}
In real implementation, it is impossible to achieve the rendering procedure defined in Eq.~\ref{eq: render} by traversing all the points along the ray. Therefore, samples carrying useful information need to be obtained for rendering. As shown in Fig.~\ref{fig: sample_comp}, NeRF~\cite{mildenhall2020nerf} and most of its variants~\cite{mildenhall2020nerf,Barron_2021_ICCV,Hedman_2021_ICCV,Reiser_2021_ICCV} adopt a hierarchical sampling strategy, which requires two field networks. In the first stage, $N_c$ coarse points are sampled by randomly drawing one from each of $N_c$ evenly-partitioned bins:
\begin{equation}
    t_i \sim \mathcal{U} [t_n + \frac{i-1}{N_c} (t_f - t_n), t_n + \frac{i}{N_c} (t_f - t_n)],
\end{equation}
where $t_n$ and $t_f$ denote the near and far bound respectively. The $N_c$ coarse samples are mapped by the first field network into colors and densities. In the second stage, $N_f$ fine samples are generated based on properties from $N_c$ coarse samples in the first stage. In \cite{mildenhall2020nerf}, weight $w_i = T_i(1 - \text{exp}(-\sigma_i \delta_i))$ of each coarse sample computed in Eq.~\ref{eq: render} serves as the probability for sampling fine points. Then both coarse and fine samples are fed into the second field network for final rendering.

\subsection{Neural Sample Field}
As aforementioned, color and density properties of $N_c$ coarse samples need to be first inferred by a field network to generate fine samples for final rendering. This procedure takes a large amount of computation cost, \ie $25\%$ of the total cost for 64 coarse samples and 128 fine samples. Inspired by the neural field concept as Eq.~\ref{eq: field}, we propose a neural sample field $\rmS_{\Theta}$ which maps a ray $\rvr(t) = \rvo + t\rvd$ directly into a series of samples for volume rendering:
\begin{equation}
\label{eq: smaple_field}
    {\rvx_1, \rvx_2, ..., \rvx_N} \gets \rmS_{\Theta} (\rvo, \rvd),
\end{equation}
where $\rvx_i$ denotes the coordinates of the $i$th sample, $N$ denotes the number of desired samples. Specifically, we first obtain $N$ scalars by feeding the ray origin coordinates and direction into $\rmS_{\Theta}$:
\begin{equation}
\label{eq: smaple_fuc}
    {\hat{t}_1, \hat{t}_2, ..., \hat{t}_N} = \rmS_{\Theta} (x_o, y_o, z_o, \theta, \phi),
\end{equation}
where $\hat{t}_i \sim [0, 1]$ represents the relative sample position between the near and far bound along the ray. $\hat{t}_i$ is mapped to a absolute position by computing $t_i = (1-\hat{t}_i)t_n + \hat{t}_it_f$. Then we compute the coordinates of the $i$th sample with
\begin{equation}
\label{eq: sample_coord}
    \rvx_i = \rvo + t_i \rvd.
\end{equation}

\begin{figure}[thbp]
    \centering
    \includegraphics[width=\linewidth]{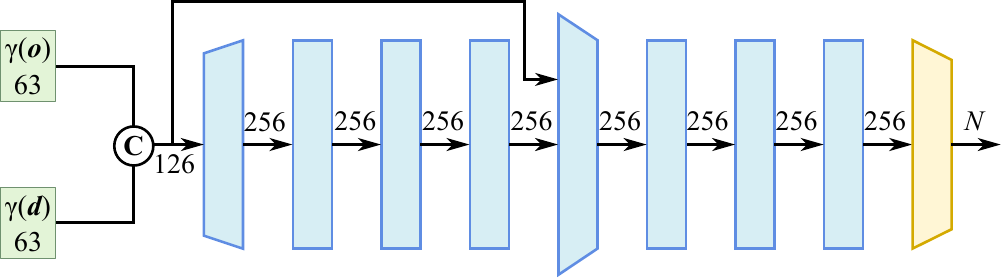}
    \caption{Architecture of the neural sample field network. ``$\gamma$'' denotes position encoding, ``\textbf{C}'' denotes concatenation. The blue blocks indicate fully connected layers followed by ReLU activation and yellow blocks indicate layers with sigmoid activation.}
    \label{fig: arch}
\end{figure}

\vspace{-18pt}
\paragraph{Architecture}
We show the architecture of the sample field network $\rmS_{\Theta}$ in Fig.~\ref{fig: arch}. The ray vector with origin coordinates and direction are first mapped into a higher dimension with position encoding, which is widely used in previous works~\cite{mildenhall2020nerf,TancikSMFRSRBN20,sitzmann2019siren,Barron_2021_ICCV}. Then the mapped input is passed through $8$ fully connected (FC) layers with ReLU activations. A skip connection is included by concatenating input with the 4-th layer's output. The hidden dimension of features between FC layers is set as 256. At the end of the network, an additional FC layer maps 256-dimension features into an $N$-dimension vector. The vector is finally fed into a sigmoid activation layer and becomes the relative sample positions $(\hat{t}_1, \hat{t}_2, ..., \hat{t}_N)$ defined in Eq.~\ref{eq: smaple_fuc}.

\vspace{-10pt}
\paragraph{Overall Framework}
As shown in Fig.~\ref{fig: framework}, we integrate the proposed neural sample fields with radiance fields as the overall framework. To render a pixel in the image, we first compute the camera pose and transform the pose to the ray origin and direction according to the pixel position. Then we feed the ray origin coordinates and normalized direction vector into the neural sample field network $\rmS_{\Theta}$. The sample field network outputs distributions within $[0, 1]$ which are transformed into 3D coordinates. Samples are then fed into the second neural radiance field to obtain corresponding colors and densities. Finally, the pixel color is generated using volume rendering as Eq.~\ref{eq: render}. Noting that both the sample field and radiance field network are optimized by minimizing the final rendered color loss as Eq.~\ref{eq: color_loss}. The whole framework can be trained end to end via gradient descent. The sample field is intuitively learning how to sample points along rays, which is intrinsically modelling geometry structures of the scene.

\subsection{Sample Field Extraction}
\label{sec: extraction}
\begin{figure}[thbp]
    \centering
    \includegraphics[width=\columnwidth]{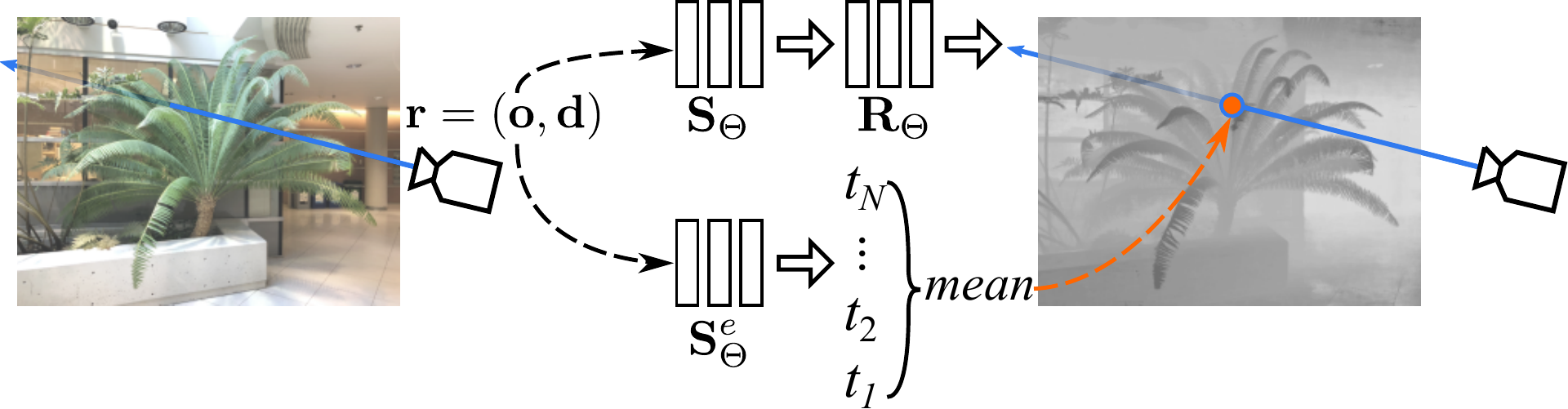}
    \vspace{-10pt}
    \caption{Depth boost for initializing an extracted sample field. The output mean value of the extracted field is forced to fit the depth predicted by a learned regular field.}
    \label{fig: extraction}
    \vspace{-10pt}
\end{figure}

Besides saving computation cost from coarse fields, we propose to extract the learned sample field for further acceleration. As shown in Fig.~\ref{fig: extraction}, we first train a \textbf{regular} sample field network $\rmS_{\Theta}$ which predicts $N$ substantial samples (\eg $N=192$ as in the fine network of NeRF~\cite{mildenhall2020nerf}) for the radiance field $\rmR_{\Theta}$ learning. Then we reduce the output number of $\rmS_{\Theta}$ to $N_e < N$ and obtain an \textbf{extracted} sample field $\rmS_{\Theta}^e$ followed by radiance field $\rmR_{\Theta}^e$. Parameters of the regular sample field $\rmS_{\Theta}$ and radiance field $\rmR_{\Theta}$ are mapped to the extracted ones $\rmS_{\Theta}^e$ and $\rmR_{\Theta}^e$. The two radiance field networks share the same architecture, so parameters from $\rmR_{\Theta}$ can be directly copied to $\rmR_{\Theta}^e$. For the sample fields, only the final FC layers for sample prediction differ where parameters are evenly mapped from $\rmS_{\Theta}$ to $\rmS_{\Theta}^e$ on the output channel dimension. All the other parameters are directly copied. With parameters mapped, we fine-tune the extracted fields $\rmS_{\Theta}^e$ and $\rmR_{\Theta}^e$ only for a few iterations to fit the new distribution.

\begin{table*}[t]
\centering
\caption{PSNR comparisons on the Realistic Synthetic $360^{\circ}$ dataset.}
\resizebox{\linewidth}{!}{
\begin{threeparttable}
\begin{tabular}{l|c|c|ccccccccc}
\toprule
\textbf{Method} & \textbf{Inf. Cost} & \textbf{Average} & Chair & Drums & Ficus & Hotdog & Lego & Materials & Mic & Ship \\
\midrule
NeRF~\cite{mildenhall2020nerf} & 1.00 & 31.01 &  33.00 & \textbf{25.01} & 30.13 & 36.18 & 32.54 & 29.62 & \textbf{32.91} & 28.65  \\
\name &  0.76  & \textbf{31.15} & \textbf{33.02} & 24.99 & \textbf{30.72} & \textbf{36.29} & \textbf{33.17} & \textbf{29.66} & 32.68 & \textbf{28.65} \\
\midrule
AutoInt~\cite{lindell2021autoint} ($N=32$) & 0.31 & 26.83 &  25.82 & 22.02 & 25.51 & 31.84 & 27.26 & \textbf{28.58} & 28.42 & 25.18\\
\name ($N_e=64$) & 0.25 & \textbf{28.39} & \textbf{29.96} & \textbf{23.43} & \textbf{27.53} & \textbf{34.41} & \textbf{29.14} & 27.76 & \textbf{29.42} & \textbf{25.47} \\
\bottomrule
\end{tabular}
\begin{tablenotes}
    \item[*] ``Inf. Cost'' denotes the relative inference cost compared with NeRF~\cite{mildenhall2020nerf}, \ie time for rendering one image which is measured on one V100 GPU. 
    \item[*] $N_e$ of \name denotes the sample number of the extracted sample field.
    \item[*] ``$N=32$'' for AutoInt~\cite{lindell2021autoint} denotes the number of piecewise sections.
\end{tablenotes}
\end{threeparttable}}
\label{tab: blender_results}
\end{table*}

For real-world scenes which usually have complicated geometry structures or depth distribution, we use the depth information predict by the regular fields to help initialize the extracted sampling filed network. We name this procedure as \textbf{depth boost}. Specifically, we sample some camera poses of the scene and feed the corresponding rays to the regular fields. For a ray $\rvr(t) = \rvo + t\rvd$, the depth $d_{\rvr}$ is predicted as:
\begin{equation}
\begin{aligned}
    {t_1, t_2, ..., t_N} &= \rmS_{\Theta} (\rvo, \rvd),\\
    \sigma_1, \sigma_2, ..., \sigma_N &= \rmR_{\Theta} (t_1, t_2, ..., t_N),\\
    d_{\rvr} &= \sum_{i=1}^N T_i(1 - \text{exp}(-\sigma_i \delta_i))t_i.
\end{aligned}
\end{equation}
We make the mean value of the extracted sample field output fit the predicted depth value $d_{\rvr}$ by minimizing the loss:
\begin{equation}
\begin{aligned}
    &{t_1^e, t_2^e, ..., t_N^e} = \rmS_{\Theta}^e (\rvo, \rvd),\\
    &\mathcal{L}_d = |(t_1^e + t_2^e + ... + t_N^e) / N^e - d_{\rvr}|.
\end{aligned}
\end{equation}
Depth boost helps sample fields with fewer points quickly converge to positions with useful information along the ray. Noting that this procedure only requires gradients for the extracted sample field and is not used in subsequent fine-tuning, which is efficient and can be finished with negligible cost.

\begin{table*}[t]
\centering
\caption{PSNR comparisons on the Real Forward-Facing dataset.}
\resizebox{\linewidth}{!}{
\begin{threeparttable}
\begin{tabular}{l|c|c|ccccccccc}
\toprule
\textbf{Method} & \textbf{Inf. Cost} & \textbf{Average} & Fern & Flower & Fortress & Horns & Leaves & Orchids & Room & T-Rex \\
\midrule
\multicolumn{11}{l}{\textbf{1008 $\times$ 756 Resolution}}\\
NeRF~\cite{mildenhall2020nerf} & 1.00 & 26.50 & \textbf{25.17} & 27.40 & 31.16 & 27.45 & 20.92 & 20.36 & 32.70 & 26.80  \\
NeRF-ID$^{\dagger}$ & $>$1.00 &  26.76 &  25.01 & 27.85 & \textbf{31.51} & 27.88 & 21.09 & \textbf{20.38} & 32.93 & 27.45  \\
\name & 0.76 & \textbf{26.83} & 24.99 & \textbf{28.14} & 31.26 & \textbf{28.32} & \textbf{21.10} & 20.08 & \textbf{33.26} & \textbf{27.46}\\
\name ($N_e=64$) & 0.25 & 26.50 & 24.77 & 28.03 & 31.09 & 27.45 & 21.06 & 20.03 & 32.71 & 26.82\\
\midrule
\multicolumn{11}{l}{\textbf{504 $\times$ 378 Resolution}}\\
NeRF~\cite{mildenhall2020nerf} & 1.00 & 27.93 & \textbf{26.92} & \textbf{28.57} & \textbf{32.94} & 29.26 & 22.50 & \textbf{21.37} & 33.60 & 28.26 \\
\name & 0.76 & \textbf{28.14} & 26.84 & 28.36 & 32.76 & \textbf{30.20} & \textbf{22.50} & 20.99 & \textbf{34.22} & \textbf{29.23} \\
\midrule
AutoInt~\cite{lindell2021autoint} ($N=32$) & 0.31 & 25.53 & 23.51 & 28.11 & 28.95 & 27.64 & 20.84 & 17.30 & 30.72 & 27.18 \\
\name ($N_e=64$) & 0.25 & \textbf{27.80} & \textbf{26.74} & \textbf{28.24} & \textbf{32.33} & \textbf{29.37} & \textbf{22.45} & \textbf{20.89} & \textbf{33.61} & \textbf{28.79} \\
\name ($N_e=32$) & 0.13 & 26.94 & 26.24 & 27.95 & 31.87 & 27.48 & 22.33 & 20.50 & 31.56 & 27.55 \\
\bottomrule
\end{tabular}
\begin{tablenotes}
    \item[*] $\dagger$ denotes training with a larger batch size, \ie 66k in NeRF-ID~\cite{arandjelovic2021nerf}.
    \item[*] Inference cost for NeRF-ID includes the additional proposal network.
\end{tablenotes}
\end{threeparttable}
}\vspace{-5pt}
\label{tab: llff_results}
\end{table*}

\begin{table*}[t]
\centering
\small
\caption{Comparison with state-of-the-art methods on Real Forward-Facing scenes.}
\vspace{-10pt}
\begin{threeparttable}
\setlength{\tabcolsep}{1mm}{
\begin{tabular}{rcccccccccc} 
\toprule
& SRN~\cite{sitzmann2019scene} & LLFF~\cite{mildenhall2019llff} & NeRF~\cite{mildenhall2020nerf} & DeRF~\cite{Rebain_2021_CVPR} & IBRNet~\cite{wang2021ibrnet} & GRF~\cite{trevithick2021grf} & SNeRG~\cite{Hedman_2021_ICCV} & NeRF-ID~\cite{arandjelovic2021nerf} & \name \\
\midrule
\textbf{PSNR} $\uparrow$ & 22.84 & 24.13 & 26.50 & 24.81 & 26.73 & 26.64 & 25.63 & 26.76 & \textbf{26.83} \\
\textbf{SSIM} $\uparrow$ & 0.668 & 0.798 & 0.811 & 0.767 & \textbf{0.851} & 0.837 & 0.818 & 0.822 & 0.823 \\
\textbf{LPIPS} $\downarrow$ & 0.378 & 0.212 & 0.250 & 0.274 & \textbf{0.175} & 0.178 & 0.183 & $\backslash$ & 0.231 \\
\bottomrule
\end{tabular}}
\end{threeparttable}
\label{tab: llff_sota}
\end{table*}

\begin{figure*}[thbp]
    \centering
    \includegraphics[width=\linewidth]{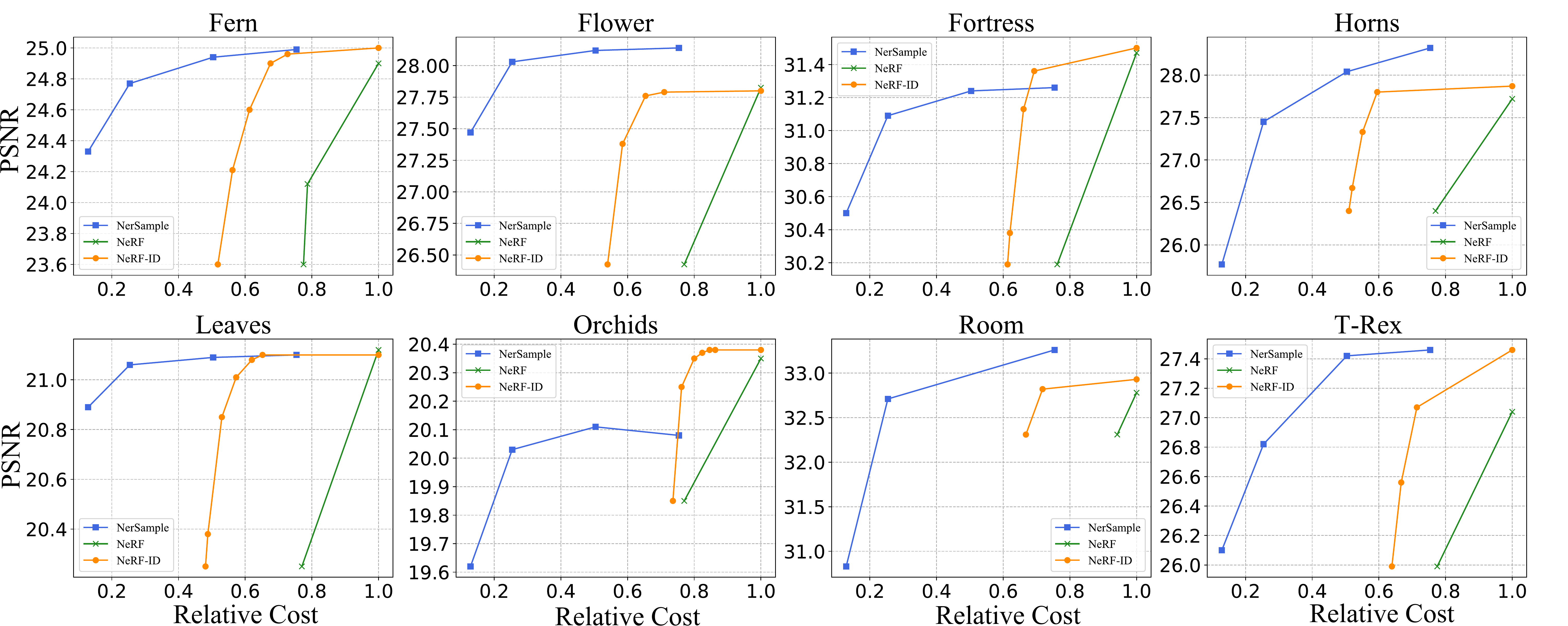}
    \vspace{-10pt}
    \caption{Speedup comparisons on the Real Forward-Facing dataset with NeRF~\cite{mildenhall2020nerf} and NeRF-ID~\cite{arandjelovic2021nerf}.}\vspace{-10pt}
    \label{fig: speedup_comp}
\end{figure*}

\section{Experiments}
\label{sec: relate}
In this section, we first describe the implementation and experimental details (Sec.~\ref{sec: implem}). Then we show results on two widely used benchmarks, \ie Realistic Synthetic $360^{\circ}$ and Real Forward-Facing datasets, and compare with other methods (Sec.~\ref{sec: results}). Some ablation studies are performed and shown in Sec.~\ref{sec: ablat}.

\subsection{Implementation Details}
\label{sec: implem}
\paragraph{Architecture Settings} Our framework is implemented based on PyTorch. We use the same network architecture for all the evaluated scenes. The sample field network is set as Fig.~\ref{fig: arch} depicts and the radiance field network is used as the same one in NeRF~\cite{mildenhall2020nerf}. For the sample field input, we apply 10-frequency position encoding to both ray origin $\rvo$ and direction $\rvd$, and the same position encoding as NeRF for radiance field input. The output number of the sample field is set as $192$ without specified in the following part.

\vspace{-10pt}
\paragraph{Training Hyperparameters} We use the Adam optimizer with a batch size of $4,096$ rays. The learning rate decays from $5\times10^{-4}$ to $5\times10^{-6}$ following a polynomial strategy with power $1$. For the Realistic Synthetic $\mathbf{360^{\circ}}$ dataset, we train each scene for $400k$ iterations in total. For the Real Forward-Facing scenes, we train each one for $100$ epochs, where each epoch is completed by randomly sampling $4,096$ rays from the whole training set for each iteration. Random noise with $0$ mean and unit variance is added to the radiance field's output densities as NeRF. 

\vspace{-10pt}
\paragraph{Extraction Hyperparameters}
For Realistic Synthetic $360^{\circ}$ scenes, the geometry structure is not complicated so we directly map parameters of regular fields to the extracted ones without depth boost (which we find no additional gain in experiments). For Real Forward-Facing scenes, we sample $120$ camera poses with a spiral path to perform depth boost. $8,192$ rays are randomly sampled in each iteration. This procedure takes only one epoch with a $5\times10^{-5}$ learning rate. The subsequent fine-tuning takes $40k$ iterations for synthetic scenes and $20$ epochs for real scenes.

\vspace{-10pt}
\paragraph{Datasets} We use two datasets, Realistic Synthetic $360^{\circ}$ and Real Forward-Facing, to evaluate our method, which are also used in NeRF~\cite{mildenhall2020nerf} and most subsequent related methods~\cite{Barron_2021_ICCV,lindell2021autoint,Reiser_2021_ICCV,wizadwongsa2021nex,arandjelovic2021nerf,Hedman_2021_ICCV}. The Realistic Synthetic $360^{\circ}$ dataset consists of $8$ scenes and each one includes 100 views for training and 200 views for testing. We take all views with an $800\times800$ resolution. The Real Forward-Facing dataset contains $8$ complex real-world scenes from NeRF~\cite{mildenhall2020nerf} and LLFF~\cite{mildenhall2019llff}. Each scene includes $20$ - $62$ images. Following \cite{mildenhall2020nerf}, we hold out $\frac{1}{8}$ images for testing and the rest are for training. All images are at $1008\times756$ pixels for experiments if unspecified.

\subsection{Results and Comparisons}
\label{sec: results}
\begin{figure*}[t!]
    \centering
    \includegraphics[width=0.95\linewidth]{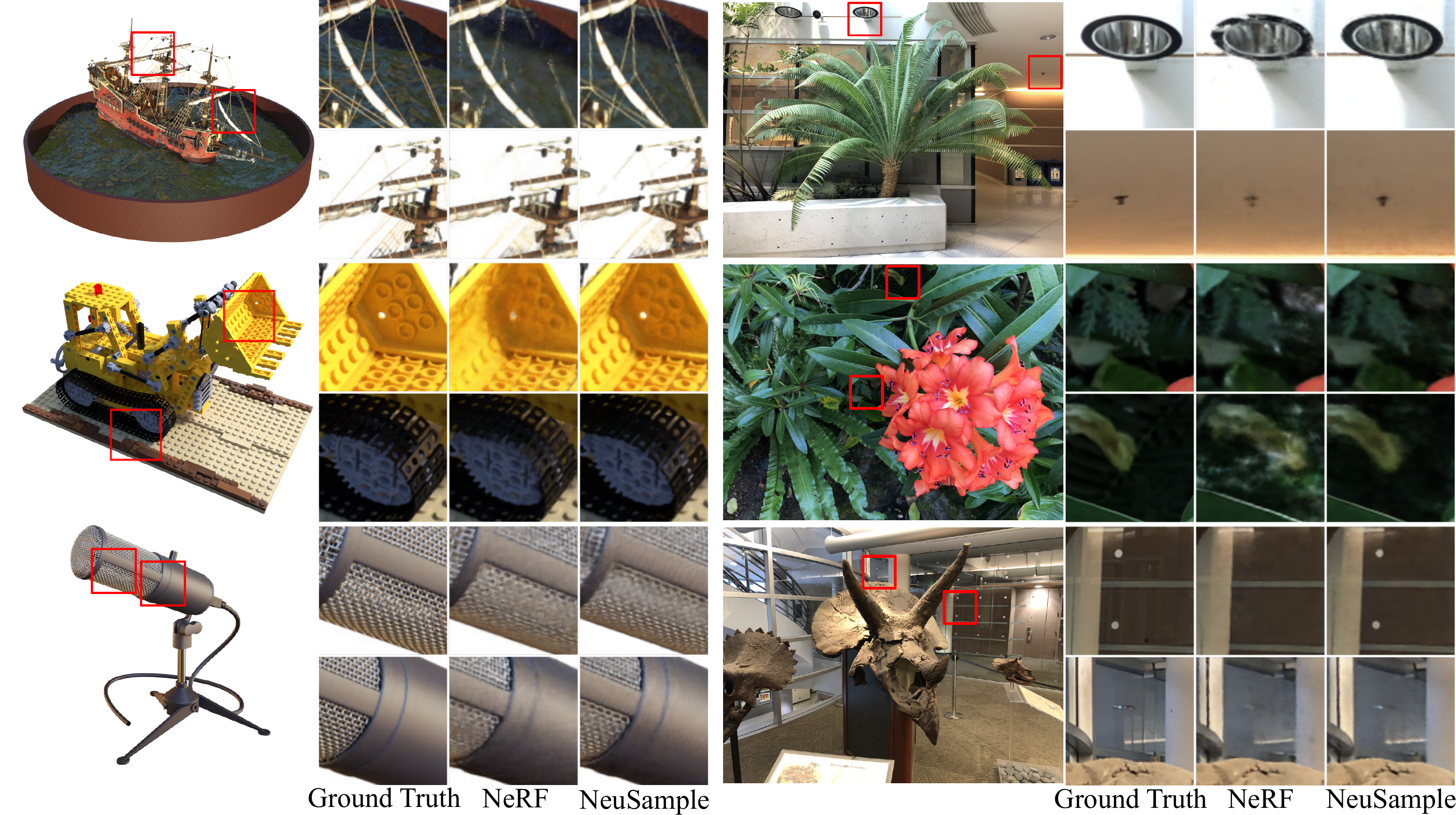}
    \vspace{-4pt}
    \caption{Visualization comparisons on the Realistic Synthetic $360^{\circ}$ (\emph{left}) and Real Forward-Facing (\emph{right}) dataset. Our method shows higher quality on rendering detailed of thin structures in the scene.}
    \vspace{-10pt}
    \label{fig: qua_vis}
\end{figure*}

\paragraph{Realistic Synthetic 360$^{\circ}$}
We show main PSNR results in Tab.~\ref{tab: blender_results} and compare with NeRF~\cite{mildenhall2020nerf} and AutoInt~\cite{lindell2021autoint}, which is a very related work focusing on reducing inference cost. Though $\sim 25\%$ computation cost of course network inference is saved, our \name still achieves similar or better PSNR performance compared with NeRF~\cite{mildenhall2020nerf}, \eg $+0.59$ PSNR for the Ficus scene and $+0.63$ for Lego. When the sample field is extracted to output $64$ points, \name achieves significantly better performance than AutoInt, \ie $1.56$ better average PSNR. The qualitative visualization results are shown in Fig.~\ref{fig: qua_vis}. Noting that though NeRF has produced high-quality synthesis results, \name shows advantages in modelling some details.

\vspace{-8pt}
\paragraph{Real Forward-Facing}
The Real Forward-Facing dataset is more challenging as it contains complex scenes in real-world scenarios, which requires more elaborate geometry structure modelling than synthetic ones. As shown in Tab.~\ref{tab: llff_results}, we perform experiments on two resolutions, \ie $1008 \times 756$ and $504 \times 378$, for better comparisons. Under the large resolution, our method achieves a high average PSNR compared with both NeRF and NeRF-ID\footnote{NeRF-ID uses a very large batch size, \ie 66k, on 16 Cloud TPUs. This training strategy can significantly promotes the baseline PSNR by $0.82$ on synthetic datasets and $0.16$ on real datasets, which we find it hard to reproduce. Therefore we only compare with NeRF-ID on Real Forward-Facing scenes for fairness.}. When we reduce the sample number to $64$, \name still achieves $26.50$ PSNR as high as NeRF while the computation cost has been decreased to $~25.4\%$ of NeRF. For the half resolution setting, \name shows $0.21$ higher PSNR than NeRF. With fewer samples, \name consistently outperforms AutoInt for all scenes by a large margin, $+2.27$ PNSR for 64 points $+1.41$ for 32 points. We comprehensively compare with other state-of-the-art methods on the Real Forward-Facing dataset in Tab.~\ref{tab: llff_sota}. \name still achieves competitive rendering quality. The qualitative results are provided in Fig.~\ref{fig: qua_vis} and \name performs better in detail modelling as well. 

\vspace{-8pt}
\paragraph{Evaluation with Diverse Sample Numbers}
To comprehensively compare with the state-of-the-art method NeRF-ID~\cite{arandjelovic2021nerf}, we evaluate \name with diverse output numbers of the sample field, \ie 192, 128, 64 and 32, which reduce the computation cost of NeRF to 75.4\%, 50.4\%, 25.4\% and 12.9\% respectively. As shown in Fig.~\ref{fig: speedup_comp}, \name achieves evidently better trade-off between rendering quality and computation cost than both NeRF and NeRF-ID for almost all scenes. 

The above experiments demonstrate that the proposed sample field can not only save cost from course network inference, but also capture more elaborate samples for learning scenes and rendering images. Even though the sample field is extracted, the radiance field can still render high-quality images with samples which carry useful information.

\subsection{Ablation Study}
\label{sec: ablat}
We perform ablation studies on two scenes, \ie, Realistic Synthetic $360^{\circ}$ lego at $800 \times 800$ and Real Forward-Facing fern at $1008 \times 756$.

\begin{table}[thbp]
\newcommand{\xmark}{\ding{55}}
\centering
\caption{Depth boost effectiveness study on the ``fern'' scene of the Real Forward-Facing dataset.}
\small
\begin{tabular}{c|c|c|c|c}
\toprule
\textbf{\#Sample} & \textbf{Depth Boost} & \textbf{PSNR} $\uparrow$ & \textbf{SSIM} $\uparrow$ & \textbf{LPIPS} $\downarrow$ \\
\midrule
128 & - & 24.99 & 0.798 & 0.272 \\
\midrule
\multirow{2}{*}{64} & \checkmark & 24.77 & 0.785 & 0.289 \\
& \xmark & 24.65 & 0.782 & 0.290 \\
\midrule
\multirow{2}{*}{32} & \checkmark & 24.33 & 0.765 & 0.307 \\
& \xmark & 24.05 & 0.753 & 0.318 \\
\bottomrule
\end{tabular}\vspace{-10pt}
\label{tab: abla_dpboost}
\end{table}

\begin{figure}[thbp]
    \centering
    \includegraphics[width=\linewidth]{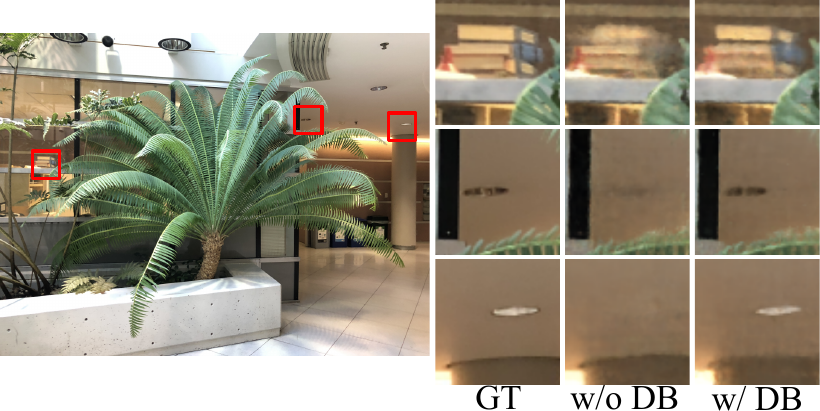}
    \vspace{-10pt}
    \caption{Rendering quality comparisons of sample field extraction \emph{w/o} and \emph{w/} depth boost on the $1008 \times 756$ fern scene. ``GT'' denotes ground truth and ``DB'' denotes depth boost. The extracted field without depth boost fails to render some small objects in the scene.}\vspace{-10pt}
    \label{fig: db_comp}
\end{figure}

\vspace{-12pt}
\paragraph{Depth Boost for Field Extraction}
We evaluate effectiveness of depth boost proposed in Sec.~\ref{sec: extraction}. As shown in Tab.~\ref{tab: abla_dpboost}, we perform this ablation study on two sample numbers, \ie $64$ and $32$. For the 64-sample setting, extracting the sample field without depth boost leads to $0.12$ PSNR decay. When the samples become fewer to $32$, PSNR degrades more by $0.28$. We show the rendering results in Fig.~\ref{fig: db_comp} and find without depth boost, some small objects in the scene are omitted while the field with depth boost models accurate outline of these objects. These defects though cause small changes to PSNR values but are evident in the final rendered image. This experiment reveals that a few iterations of depth boost to initialize the sample field effectively helps to locate points with high importance. Even with few samples, the real pixel color can be rendered accurately.

\begin{table}[t!]
\centering
\caption{Layer number study of the sample field network on the synthetic lego and real fern scenes.}
\small
\begin{tabular}{l|c|c|c|cccccccc}
\toprule
\textbf{Scene} & \textbf{\#Layers} & \textbf{PSNR} $\uparrow$ & \textbf{SSIM} $\uparrow$ & \textbf{LPIPS} $\downarrow$ \\
\midrule
\multirow{3}{*}{lego} & 8 & 33.17 & 0.965 & 0.048 \\
& 4 & 29.29 & 0.934 & 0.106 \\
& 2 & 28.81 & 0.937 & 0.090 \\
\midrule
\multirow{3}{*}{fern} & 8 & 24.99 & 0.798 & 0.272 \\
& 4 & 24.70 & 0.778 & 0.303 \\
& 2 & 24.79 & 0.790 & 0.283 \\
\bottomrule
\end{tabular}\vspace{-10pt}
\label{tab: abla_layers}
\end{table}

\vspace{-10pt}
\paragraph{Layer Numbers of Sample Field Network}
We study the layer number design for constructing the sample field network. As shown Tab.~\ref{tab: abla_layers}, three layer settings are evaluated, \ie 8 (the default setting), 4 and 2, on the two scenes of lego and fern. We observe that the layer number decrease causes slight impact for the fern scene rendering within $\sim 0.29$ PSNR decay. However, smaller layer numbers lead to drastically rendering quality degradation for the lego scene. Setting to 4 layers drops PSNR by $3.88$; and setting to 2 causes $4.36$ decay. We analyze this phenomenon and deduce the reason as follows. The synthetic lego scene contains more views in a larger range than fern, so it requires a deeper neural network with more parameters to fit diverse view-dependent sampling distributions. 

\begin{table}[thbp]
\centering
\caption{Frequency study of position encoding on the synthetic lego and real fern scenes.}
\small
\begin{tabular}{l|c|c|c|cccccccc}
\toprule
\textbf{Scene} & \textbf{\#Frequencies} & \textbf{PSNR} $\uparrow$ & \textbf{SSIM} $\uparrow$ & \textbf{LPIPS} $\downarrow$ \\
\midrule
\multirow{3}{*}{lego} & 5 & 33.09 & 0.966 & 0.047 \\
& 10 & 33.17 & 0.965 & 0.048 \\
& 15 & 33.02 & 0.965 & 0.047 \\
\midrule
\multirow{3}{*}{fern} & 5 & 24.80 & 0.790 & 0.283 \\
& 10 & 24.99 & 0.798 & 0.272 \\
& 15 & 25.12 & 0.800 & 0.270 \\
\bottomrule
\end{tabular}
\label{tab: abla_freqs}
\vspace{-10pt}
\end{table}

\begin{figure}[thbp]
    \centering
    \includegraphics[width=0.9\linewidth]{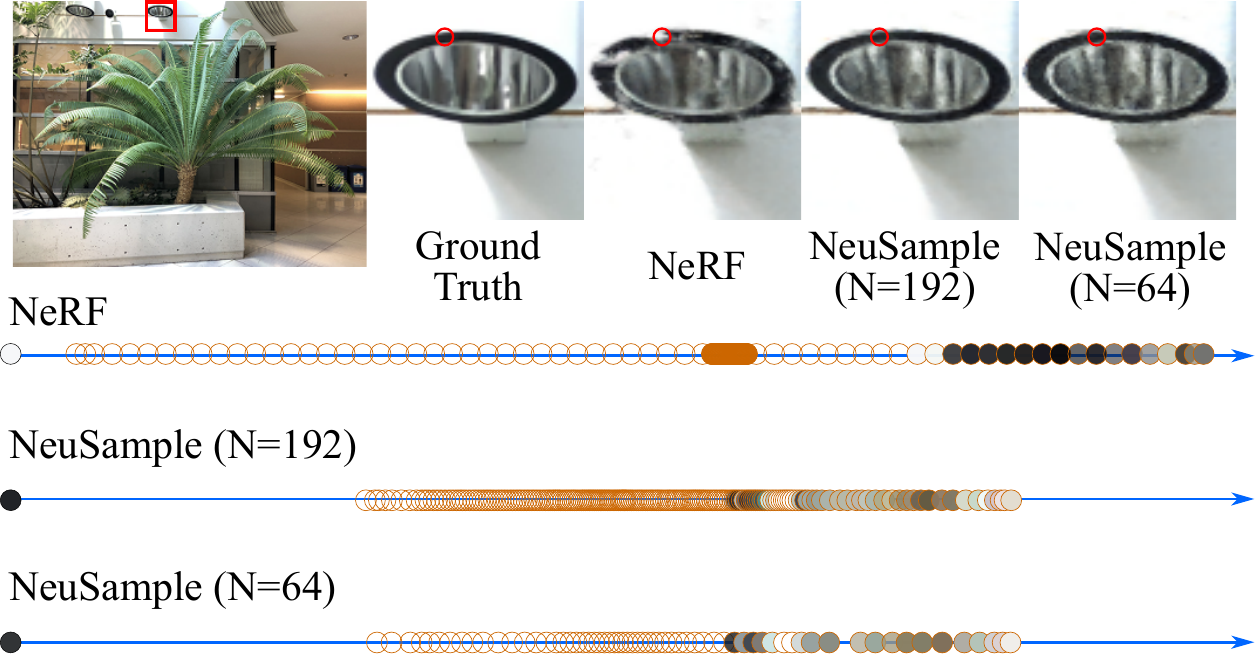}
    \caption{Sample visualization comparisons along the ray in the real scene ``fern''. Samples are colored with ``RGBA'' values, which are set as predicted colors and densities. The edge of each sample is of the orange color, while the orange mass in the NeRF ray represents that lots of points are gathered. The start point is with the finally rendered color.\vspace{-10pt}}
    \label{fig: sample_comp_fern}
\end{figure}

\vspace{-10pt}
\paragraph{Frequencies of Position Encoding}
We study frequencies of position encoding in the sample field network. As shown in Tab.~\ref{tab: abla_freqs}, we evaluate three frequency numbers 5, 10 (the default setting) and 15 on two scenes of lego and fern. We find changing frequencies in this range does not affect the sample field much, $< 0.2$ PSNR for both scenes.

\vspace{-10pt}
\paragraph{Sample Visualization}
As shown in Fig.~\ref{fig: sample_comp_fern}, we visualize the obtained samples of three models, \ie NeRF~\cite{mildenhall2020nerf}, our \name with $192$ points and an extrated \name with $64$ points. We observe NeRF tends to sample points in a wide range from the near bound to far bound but locate too many samples at a similar position (the orange mass). On the contrary, \name locates points in a narrower range but the points are more even. With fewer points, \name can still obtain the key samples which render accurate colors. We deduce that an extremely dense distribution around a local point is not beneficial for the field training, \eg NeRF predicts wrong opacity values for key points in this example.

\section{Conclusion}
\label{sec:conclu}
In this paper, we propose a neural sample field which maps a ray to sampling distributions. The proposed sample field can be integrated with neural radiance fields for volume rendering, which we name \name. Our method obtains samples for rendering with a lightweight network module, which saves computation cost from widely used course networks and shows stronger ability of synthesizing novel views for 3D scenes. With a learned \name, the sample field can be extracted for further acceleration with maintaining high rendering quality.

\vspace{-10pt}
\paragraph{Limitations} The proposed \name approach has similar limitations as radiance fields in previous methods~\cite{mildenhall2020nerf,kaizhang2020,Barron_2021_ICCV} which needs to be optimized towards every independent scene, as the sample field is directly related to the geometry structure of each specific scene. The methods for generalizing the learned field or speeding up the optimization procedure~\cite{tancik2021learned,yu2021pixelnerf,Jain_2021_ICCV,sitzmann2021lfns,trevithick2021grf} could be integrated with our method for further reducing the training cost.

Besides, though \name accelerates rendering, it is unlikely that purely using \name can achieve real-time rendering like methods with caching-based techniques~\cite{Reiser_2021_ICCV,Hedman_2021_ICCV,Yu_2021_ICCV,Garbin_2021_ICCV,wizadwongsa2021nex}. In comparison, \name and~\cite{lindell2021autoint,neff2021donerf,arandjelovic2021nerf} accelerate rendering from another path that focuses on the field itself and requires no additional storage. Overall, this is a problem of achieving better trade-off between storage and speed, and we believe that integrating the above methodologies is a promising future research direction.

\section*{Acknowledgements}
We sincerely thank Liangchen Song, Yingqing Rao and Yuzhu Sun for their generous assistance and discussion.

{\small
\bibliographystyle{ieee_fullname}
\bibliography{egbib}
}

\begin{table*}[t]
\centering
\caption{Per-scene quantitative comparisons on the Realistic Synthetic $360^{\circ}$ dataset.}
\resizebox{\linewidth}{!}{
\begin{threeparttable}
\begin{tabular}{l|c|c|ccccccccc}
\toprule
\textbf{Method} & \textbf{Inf. Cost} & \textbf{Average} & Chair & Drums & Ficus & Hotdog & Lego & Materials & Mic & Ship \\
\midrule

\multicolumn{11}{c}{\textbf{PSNR $\uparrow$}}\\
NeRF~\cite{mildenhall2020nerf} & 1.00 & 31.01 &  33.00 & \textbf{25.01} & 30.13 & 36.18 & 32.54 & 29.62 & \textbf{32.91} & 28.65  \\
\name & \textbf{0.76}  & \textbf{31.15} & \textbf{33.02} & 24.99 & \textbf{30.72} & \textbf{36.29} & \textbf{33.17} & \textbf{29.66} & 32.68 & \textbf{28.65} \\
\midrule
AutoInt~\cite{lindell2021autoint} ($N=32$) & 0.31 & 26.83 &  25.82 & 22.02 & 25.51 & 31.84 & 27.26 & \textbf{28.58} & 28.42 & 25.18\\
\name ($N_e=64$) & \textbf{0.25} & \textbf{28.39} & \textbf{29.96} & \textbf{23.43} & \textbf{27.53} & \textbf{34.41} & \textbf{29.14} & 27.76 & \textbf{29.42} & \textbf{25.47} \\
\bottomrule

\multicolumn{11}{c}{}\\
\multicolumn{11}{c}{\textbf{SSIM $\uparrow$}}\\
NeRF~\cite{mildenhall2020nerf} & 1.00 & 0.947 & 0.967 & \textbf{0.925} & 0.964 & 0.974 & 0.961 & 0.949 & 0.980 & 0.856 \\
\name &  \textbf{0.76}  & \textbf{0.949} & \textbf{0.968} & 0.924 & \textbf{0.968} & \textbf{0.977} & \textbf{0.965} & 0.949 & 0.980 & \textbf{0.863} \\
\midrule
AutoInt~\cite{lindell2021autoint} ($N=32$) & 0.31 & \textbf{0.927} & 0.926 & 0.885 & 0.926 & \textbf{0.973} & 0.929 & \textbf{0.953} & 0.951 & \textbf{0.869} \\
\name ($N_e=64$) & \textbf{0.25} & 0.923 & \textbf{0.941} & \textbf{0.897} & \textbf{0.942} & 0.966 & \textbf{0.931} & 0.927 & \textbf{0.964} & 0.819 \\
\bottomrule

\multicolumn{11}{c}{}\\
\multicolumn{11}{c}{\textbf{LPIPS $\downarrow$}}\\
NeRF~\cite{mildenhall2020nerf} & 1.00 & 0.081 & 0.046 & 0.091 & 0.044 & 0.121 & 0.050 & \textbf{0.063} & 0.028 & 0.206 \\
\name &  \textbf{0.76}  & \textbf{0.068} & \textbf{0.045} & 0.091 & \textbf{0.036} & \textbf{0.043} & \textbf{0.048} & 0.073 & \textbf{0.027} & \textbf{0.183} \\
\midrule
AutoInt~\cite{lindell2021autoint} ($N=32$) & 0.31 & 0.152 & 0.149 & 0.209 & 0.109 & 0.088 & 0.135 & 0.100 & 0.127 & 0.295 \\
\name ($N_e=64$) & \textbf{0.25} & \textbf{0.105} & \textbf{0.077} & \textbf{0.133} & \textbf{0.073} & \textbf{0.067} & \textbf{0.097} & \textbf{0.095} & \textbf{0.057} & \textbf{0.238} \\
\bottomrule
\end{tabular}
\begin{tablenotes}
    \item[*] ``Inf. Cost'' denotes the relative inference cost compared with NeRF~\cite{mildenhall2020nerf}, \ie time for rendering one image which is measured on one V100 GPU. 
    \item[*] $N_e$ of \name denotes the sample number of the extracted sample field.
    \item[*] ``$N=32$'' for AutoInt~\cite{lindell2021autoint} denotes the number of piecewise sections.
\end{tablenotes}
\end{threeparttable}}
\label{tab: blender_quant}
\end{table*}

\begin{table*}[t]
\centering
\caption{Per-scene quantitative comparisons on the Real Forward-Facing dataset.}
\resizebox{\linewidth}{!}{
\begin{threeparttable}
\begin{tabular}{l|c|c|ccccccccc}
\toprule
\textbf{Method} & \textbf{Inf. Cost} & \textbf{Average} & Fern & Flower & Fortress & Horns & Leaves & Orchids & Room & T-Rex \\
\midrule

\multicolumn{11}{c}{\textbf{PSNR $\uparrow$}}\\
NeRF~\cite{mildenhall2020nerf} & 1.00 & 26.50 & \textbf{25.17} & 27.40 & 31.16 & 27.45 & 20.92 & 20.36 & 32.70 & 26.80  \\
NeRF-ID$^{\dagger}$~\cite{arandjelovic2021nerf} & $>$1.00$^\ddagger$ &  26.76 &  25.01 & 27.85 & \textbf{31.51} & 27.88 & 21.09 & \textbf{20.38} & 32.93 & 27.45  \\
\name & 0.76 & \textbf{26.83} & 24.99 & \textbf{28.14} & 31.26 & \textbf{28.32} & \textbf{21.10} & 20.08 & \textbf{33.26} & \textbf{27.46}\\
\name ($N_e=64$) & \textbf{0.25} & 26.50 & 24.77 & 28.03 & 31.09 & 27.45 & 21.06 & 20.03 & 32.71 & 26.82\\
\bottomrule

\multicolumn{11}{c}{}\\
\multicolumn{11}{c}{\textbf{SSIM $\uparrow$}}\\
NeRF~\cite{mildenhall2020nerf} & 1.00 & 0.811 & 0.792 & 0.827 & 0.881 & 0.828 & 0.690 & \textbf{0.641} & 0.948 & 0.880 \\
NeRF-ID$^{\dagger}$~\cite{arandjelovic2021nerf} & $>$1.00$^\ddagger$ & 0.822 & \textbf{0.800} & 0.840 & \textbf{0.890} & 0.840 & \textbf{0.710} & 0.640 & 0.950 & \textbf{0.900} \\
\name & 0.76 & \textbf{0.823} & 0.798 & \textbf{0.845} & 0.888 & \textbf{0.859} & 0.708 & 0.630 & \textbf{0.958} & 0.899 \\
\name ($N_e=64$) & \textbf{0.25} & 0.811 & 0.785 & 0.840 & 0.882 & 0.826 & 0.702 & 0.623 & 0.950 & 0.882 \\
\bottomrule

\multicolumn{11}{c}{}\\
\multicolumn{11}{c}{\textbf{LPIPS $\downarrow$}}\\
NeRF~\cite{mildenhall2020nerf} & 1.00 & 0.250 & 0.280 & 0.219 & 0.171 & 0.268 & 0.316 & \textbf{0.321} & 0.178 & 0.249 \\
\name & 0.76 & \textbf{0.231} & \textbf{0.272} & \textbf{0.192} & \textbf{0.158} & \textbf{0.218} & \textbf{0.296} & 0.331 & \textbf{0.154} & \textbf{0.224} \\
\name ($N_e=64$) & \textbf{0.25} & 0.251 & 0.289 & 0.201 & 0.167 & 0.261 & 0.306 & 0.361 & 0.173 & 0.248 \\
\bottomrule

\end{tabular}
\begin{tablenotes}
    \item[$\dagger$] denotes training with a larger batch size, \ie 66k in NeRF-ID~\cite{arandjelovic2021nerf}.
    \item[$\ddagger$] Inference cost for NeRF-ID includes the additional proposal network.
\end{tablenotes}
\end{threeparttable}
}
\label{tab: llff_quant}
\end{table*}

\appendix
\section{Appendix}
\subsection{Detailed Quantitative Results}
As shown Tab.~\ref{tab: blender_quant} and Tab.~\ref{tab: llff_quant}, we provide additional quantitative results for both Realistic Synthetic $360^{\circ}$ scenes at $800\times800$ and Real Forward-Facing scenes at $1008 \times 756$ with three evaluation metrics of PSNR, SSIM and LPIPS. Our \name consistently outperforms NeRF~\cite{mildenhall2020nerf} with only $76\%$ computation cost. With fewer samples for rendering, \name still shows promising rendering quality with a significantly better trade-off between computation budget and quality than compared methods~\cite{arandjelovic2021nerf,lindell2021autoint}.

\end{document}